\documentclass[11pt,a4paper]{article}
\usepackage[hyperref]{acl2020}
\usepackage{times}
\usepackage{latexsym}
\usepackage{graphicx}

\usepackage{amsmath}
\usepackage{amssymb}
\usepackage{multirow}
\usepackage[table,xcdraw]{xcolor}
\usepackage{tabularx}
\usepackage{makecell}
\usepackage{microtype}

\aclfinalcopy 

\newcommand{\code}[1]{\texttt{#1}}

\title{Universal Adversarial Triggers}

\author{Benedict Florance Arockiaraj \And Alexander Feng \AND
  Jianxiong Cai \And Xiaoyu Cheng}

\date{}

\begin{document}
\maketitle
\begin{abstract}
Recent works have illustrated that modern NLP models trained for diverse tasks ranging from sentiment analysis to language generation succumb to \textit{universal adversarial attacks}, a class of input-agnostic attacks where a common trigger sequence is used to attack the model. Although these attacks are successful, the triggers generated by such attacks are ungrammatical and unnatural. Our work proposes a novel technique combining parts-of-speech filtering and perplexity based loss function to generate sensible triggers that are closer to natural phrases. For the task of sentiment analysis on the SST dataset, the method produces sensible triggers that achieve accuracies as low as \textbf{0.04} and \textbf{0.12} for flipping positive to negative predictions and vice-versa. To build robust models, we also perform adversarial training using the generated triggers that increases the accuracy of the model from \textbf{0.12} to \textbf{0.48}. We aim to illustrate that adversarial attacks can be made difficult to detect by generating sensible triggers, and to facilitate robust model development through relevant defenses.

\end{abstract}

\section{Introduction}

Modern adversarial attacks are successful in attacking NLP models for diverse tasks like sentiment analysis, natural language inference (NLI), language generation, question answering etc. A particular class of attacks, called \textit{universal adversarial attacks} perform attacks using a single trigger sequence that can be applied to \textit{any} input to the desired model. Being input-agnostic, it poses a serious threat as we do not have to generate multiple triggers for every input. Although such attacks are successful, past works in this domain produce ungrammatical and irregular trigger sequences like \textit{`zoning tapping fiennes'} \cite{wallace-etal-2019-universal}, that can be easily detected using simple heuristics. 

\subsection{Task Description}
To counter this shortcoming, we aim to generate \textit{sensible, natural} trigger sequences that are input agnostic. In addition, we explore the realm of building robust models by retraining the models adversarially using the triggers generated by our trigger generation methodology. 

Our problem is centered around key areas of NLP/Computational Linguistics like adversarial attacks, adversarial defense, part-of-speech tagging and perplexity. Although we illustrate all experiments for the task of sentiment analysis, it can be extrapolated to other tasks like language generation, NLI and question-answering.

\subsection{Formal Definition}
Given a dataset, model, and a task (like sentiment analysis, language generation etc.), we intend to find \textit{sensible} universal triggers (that are agnostic to the input) and cause the model to produce a certain desired output. To put it formally, we intend to find the trigger $t$,
\begin{equation}
    \operatorname*{arg\,min}_t {\mathbb{E}}_{x\in D} L(f(x;t), \hat{y})
\end{equation}
where $f$ is a given model, $D$ is the dataset, $x$ is an example from the dataset, $\hat{y}$ is the desired output. `;' denotes concatenation of 
$x$ and $t$ ($t$ can either be appended/prepended to $x$). 

\subsection{Motivation}
Adversarial examples can even cause state-of-the-art models to go haywire and expose the model vulnerabilities. In fact, it helps us to make models more explainable and interpretable. Wouldn’t it be easier to use a common, universal trigger sequence that is agnostic of the input to attack models and achieve our desired goal? It definitely would be, and such attacks are dangerous given that the trigger sequence can be distributed widely and anyone can attack using those sequences. Our motivation stems from being interested in generating such \textit{universal and natural} trigger sequences and analyzing how state-of-the-art models in tasks like sentiment analysis succumb to such sequences. Prior works show that such adversarial triggers could be as egregious as producing offensive and racist content in language generation by exploiting bias from datasets. Tackling this serious problem, our project work aims to delve into this essential field at the confluence of adversarial attacks, bias and explainability, and contribute novel techniques by extending prior work in this domain.
\begin{table*}
\centering

\begin{tabular}{lll}
\hline
Task                                 & Input                                                           & Model Prediction                     \\ \hline
                                     &{\color[HTML]{FE0000} \textbf{zoning tapping fiennes}} Visually imaginative, thematically & Positive --\textgreater Negative     \\
                                     & instructive and thoroughly delightful, it takes \\
                                     & us on a roller-coaster ride. . . \\         
 \multirow{-2}{*}{\thead{Sentiment\\ Analysis}} & {\color[HTML]{FE0000} \textbf{irredeemably disgusting garbage}} As surreal as & Positive --\textgreater Negative     \\ 
                                     & a dream and as detailed as a photograph,\\
                                     & as visually dexterous as it is at times imaginatively overwhelming.\\
\hline
\end{tabular}

\caption{The table shows the generated \textit{universal adversarial triggers} red and the input data in black. These triggers, when appended to any input, can cause the prediction to flip for tasks like sentiment analysis. The first example is a trigger from \cite{wallace-etal-2019-universal}, while the second example is a trigger generated from our proposed method.}
\end{table*}
\section{Literature Review}
There is a sizeable body of related research for our task and extensions. We base our project on \cite{wallace-etal-2019-universal}, and we cover several other papers related to generating sensible adversarial examples and training models using adversarial inputs. 

\subsection{Universal Adversarial Triggers for Attacking and Analyzing NLP \cite{wallace-etal-2019-universal}}
This paper proposes a gradient-based method to find universal adversarial triggers that leads to malicious behavior in CNN-based models for various tasks (sentiment analysis, natural language inference, Q\&A, and GPT-2). Universal adversarial triggers are short input-agnostic phrases that trigger a model to produce a specific prediction (usually a malicious one). For example, by prepending a single-word trigger to original input sequences, a sentiment analysis model would classify a negative sample as positive. Given a white-box CNN-based model, it iteratively updates the universal trigger by backpropagating the output loss with respect to the input until it finds a good trigger (e.g., the output loss for the classification task would be cross-entropy loss between the model prediction and target malicious label). One major drawback is for multi-word triggers, the algorithm updates individual tokens separately, which leads to nonsensical input phrases. Successive works like \cite{song-etal-2021-universal} improve on that by using an adversarially regularized autoencoder (ARAE) to generate triggers jointly. 

\subsection{Semantically Equivalent Adversarial Rules for Debugging {NLP} models \cite{ribeiro-etal-2018-semantically}}
The paper attempts to reveal the oversensitivity in NLP models by presenting Semantically Equivalent Adversaries (SEAs) -  semantic-preserving perturbations that change the model prediction - and Semantically Equivalent Adversarial Rules (SEARs) - generalized replacement rules for inducing SEAs. To find SEAs, for every correctly predicted instance, the authors generate a set of paraphrases around the sentence using beam search until the model prediction changes or the Semantic Similarity Score, a measure of similarity between a paraphrase and the original sentence, drops below a threshold. To generate SEARs, the algorithm first generates a set of candidate rules from every SEA by extracting contiguous raw text or Part-of-Speech sequences that paraphrase the sentence (e.g. what NOUN $\rightarrow$ which NOUN), then filters out non-semantically-equivalent rules, then greedily selects rules that maximize submodular coverage while avoiding redundancy. Finally, a human is introduced in the loop to select the rules they think are bugs. Such generated SEARs satisfy three properties: 1) semantic equivalence, 2) high adversary count (to induce as many SEAs as possible), and 3) non-redundancy (to cover as many instances as possible without repetition). Experiments on tasks including Machine Comprehension, Visual QA, and Sentiment Analysis show the method is effective. SEARs are also actionable through data augmentation. 

\subsection{Contextualized Perturbation for Textual Adversarial Attack \cite{li-etal-2021-contextualized}}
The paper presents CLARE, a ContextuaLized AdversaRial Example generation model for text with an aim to produce fluent and grammatical triggers unlike approaches like \cite{wallace-etal-2019-universal}. Their approach uses a mask-then-infill procedure that first detects model vulnerabilities, adds a mask to the inputs to indicate missing word(s), and plugs in an alternative word(s) using a pretrained language model RoBERTa. The plugging of the alternative word is modelled in three ways: i) Replace (replaces a token), ii) Insert (inserts a new token) and iii) Merge (merges a bigram), thus producing output of varied lengths. CLARE iteratively applies a sequence of contextualized perturbation actions to the input such that the alternative trigger attains high probability from a masked language model, new sentence (with alternative trigger) looks similar to the old sentence, and our intended task model predicts low probability for the gold label given the new sentence. They perform experiments in text classification, natural language inference (NLI), and sentence paraphrase tasks that show high attack success rate, fluency and grammaticality in comparison to previous SOTA methods. In addition, they also demonstrate that adversarial training helps to decrease the attack success rate.


\subsection{Inoculation by Fine-Tuning: A Method for Analyzing Challenge Datasets \cite{liu-etal-2019-inoculation}}
Challenge datasets, created with the intention of exposing brittleness in datasets and models, can provide ambiguous results when accuracy on the challenge sets is far lower than accuracy on the original test sets, failing to pinpoint the weaknesses of the model and dataset. By training Natural Language Inference and Question Answering models with some selections from the challenge sets and evaluating again, \cite{liu-etal-2019-inoculation} assess how well the models can adapt to challenge examples. For some challenge sets, such as word overlap and negation, inoculation proves effective, greatly improving accuracy on the challenge set. For others, such as spelling errors and length mismatch, performance is largely unaffected by inoculation. For numerical reasoning and with adversarial question answering, challenge set performance is increased at the cost of a drop in accuracy on the original dataset.  By using a similar approach with universal adversarial triggers, we can potentially analyze the weaknesses exploited by the triggers and determine whether they lie in the model, the data, or in some combination thereof.

\section{Experimental Design}
\subsection{Data}

\textbf{SST (Stanford Sentiment Treebank)}
Published in 2013, the SST dataset contains fine grained sentiment labels for 215,154 phrases from 11,855 movie review sentences \cite{socher-etal-2013-recursive}. Modeled as a classification problem, each phrase is labeled as either 'very negative', 'negative', 'neutral', 'positive' or 'very positive' mapped to the probabilities [0, 0.2], (0.2, 0.4], (0.4, 0.6], (0.6, 0.8], (0.8, 1.0]. Using Amazon Mechanical Turk, annotators were presented phrases in random order which were labelled using a slider containing 25 levels of sentiment. This dataset also provides a standard train-dev-test split for benchmark evaluation. Table \ref{dataset-split} gives the train-dev-test split of the SST dataset. Table \ref{dataset-examples} shows a few examples from the dataset with their corresponding labels. \\

\begin{table}
\centering
\begin{tabular}{ll}
\hline \textbf{Split} & \textbf{Count} \\ \hline
Training & 8544 \\
Development & 1101 \\
Test & 2210 \\
\hline
\end{tabular}
\caption{\label{dataset-split} SST dataset split }
\end{table}

\begin{table*}[h!]
\centering
\begin{tabular}{|p{13cm}|p{1.5cm}|}
\hline \textbf{Example} & \textbf{Label} \\ \hline
` ... the film 's considered approach to its subject matter is too calm and thoughtful for agitprop , and the thinness of its characterizations makes it a failure as straight drama . '& 0.22222 \\ \hline
`Mafia , rap stars and hood rats butt their ugly heads in a regurgitation of cinematic violence that gives brutal birth to an unlikely , but likable , hero . ' & 0.48611 \\ \hline
`, wonderfully respectful of its past and thrilling enough to make it abundantly clear that this movie phenomenon has once again reinvented itself for a new generation' & 0.81944 \\
\hline
\end{tabular}
\caption{\label{dataset-examples} Examples from the SST dataset with labels}
\end{table*}

\subsection{Evaluation Metric}

\textbf{Accuracy} To evaluate the performance of generated universal triggers, we evaluate the model accuracy on subsets of the development dataset with and without the triggers. The subsets contain only examples of the label to flip. Since there is only one true label, all elements will be either true positives or false negatives. Ideally, the trigger should lower the original model accuracy.
\begin{equation}
Accuracy = \frac{TP}{TP + FN}
\end{equation}
where TP denotes true positive count and FN denotes false negative count.

We would usually focus on two accuracy metrics:
\begin{itemize}
    \item Accuracy on original dev set (without triggers): A good defense model should not sacrifice the model accuracy for the robustness. In other words, a good robust model should first have acceptable performance on original set. 
    \item Accuracy on attack dev set (with triggers): We generate the universal trigger using algorithm in \cite{wallace-etal-2019-universal}, and prepend the trigger at the beginning of each phrase in the original dev test to create the attack dev set. This accuracy can be interpreted as the attack effectiveness or the model robustness (depending on the actual scenario). 
\end{itemize}

\subsection{Simple Baselines}
The following are the simple baselines that we set up for this task. The results for the simple baselines can be found in tables \ref{positive-baseline} and \ref{negative-baseline}.

\subsubsection{Random Attack}
In a random attack, potential triggers are randomly sampled from the vocabulary and selected based on the loss values of models evaluating on data with the triggers prepended.

\subsubsection{Nearest Neighbor Attack}
In nearest neighbors attack, we take a small step in the direction of the averaged gradient and find the nearest vector in the embedding matrix using k-d tree.

\subsubsection{Hardcoded Attack}
In the hardcoded attack, we pick an intuitive trigger sequence based on the target task and directly evaluate the model on the sequence. Triggers used in results are the ones with the greatest effect among a few ($<3$) options.
\subsubsection{Top Frequent Words (sentiment analysis only)}
For sentiment analysis, we first split the training dataset into two splits (positive / negative) based on the sentiment label of each sentence. After removing stop words and common words (e.g. movie, film), we count word frequencies in each split and visualize the top frequent words in the appendix. Then, we generate the trigger to be the top 3 frequent words for each split. (i.e. [good, funny, comedy] for positive target label, [bad, much, characters] for negative target label).

\section{Experimental Results}
\subsection{Published Baseline}

Next, we iteratively update the trigger words in order to increase the probability of the specific target prediction. For instance, a trigger for sentiment analysis is optimized to increase the probability of the negative class for various positive movie reviews. We perform the iterative updates based on the model's gradient equation from \cite{wallace-etal-2019-universal}. We achieve similar triggers and performance accuracy as mentioned in the paper for all the different tasks. Since we use the same dataset and split as of \cite{wallace-etal-2019-universal}, our results are directly comparable. The results for the published baselines can be found in tables \ref{positive-baseline} and \ref{negative-baseline}.

\begin{table*}[]
\centering
\small
\begin{tabular}{|l |c | c | c |}
\hline
\textbf{Baseline} & \textbf{Trigger Used/Generated} & \textbf{Acc w/o trig} & \textbf{Acc with trig} \\ \hline
Random Triggers & pointless sooooo lifeless & 0.909909  & 0.112612 \\ \hline
Hardcoded Trigger & bad bad bad &  0.909909 &  0.378378\\ \hline
Top Frequent Words & bad, much, characters & 0.909909  & 0.38063 \\ \hline
Nearest Neighbor Trigger & not its forefront & 0.909909   & 0.680180 \\ \hline
Universal Adversarial Trigger \cite{wallace-etal-2019-universal} & sucks lifeless lifeless  &  0.909909  & 0.085585 \\ \hline
\end{tabular}\\[1ex]
\caption{\label{positive-baseline}Baseline attacks to flip positive sentiment to negative on the SST dataset. We present accuracy before and after trigger generation, and the corresponding generated trigger. All except the last row shows results for the simple baselines, while the last row presents the results of our strong baseline \cite{wallace-etal-2019-universal}}
\end{table*}

\begin{table*}[]
\centering
\small
\begin{tabular}{|l |c | c | c |}
\hline
\textbf{Baseline} & \textbf{Trigger Used/Generated} & \textbf{Acc w/o trig} & \textbf{Acc with trig} \\ \hline
Random Triggers & captivates unforgettable sensual & 0.892523 & 0.203271 \\ \hline
Hardcoded Trigger & positive positive positive &  0.892523 &  0.47196\\ \hline
Top Frequent Words & good, funny, comedy &  0.892523 & 0.44159 \\ \hline
Nearest Neighbor Trigger & above fascinates fascinating & 0.892523  & 0.390186 \\ \hline
Universal Adversarial Trigger \cite{wallace-etal-2019-universal} & vividly thought-provoking captivating & 0.892523 & 0.093457 \\ \hline
\end{tabular} \\[1ex]
\caption{\label{negative-baseline}Baseline attacks to flip negative sentiment to positive on the SST dataset. We present accuracy before and after trigger generation, and the corresponding generated trigger. All except the last row shows results for the simple baselines, while the last row presents the results of our strong baseline \cite{wallace-etal-2019-universal}}
\end{table*}

\subsection{Extensions}
\begin{table*}[h!]
\centering
\begin{tabular}{|p{6.7cm}|p{6.3cm}|p{1.5cm}|}
\hline
\textbf{Ablation Experiments} & \textbf{Sample Triggers} & \textbf{Accuracy} \\ \hline
Random Attack w/o Beam & {uncreative grow miserably} & 0.146396\\ \hline
UAT Attack w/o Beam & {boring forges faulty} & 0.130630\\ \hline
Random Attack + POS + Beam & {rapidly inadequate spaceship} & 0.227477\\ \hline
Random Attack + Perplexity + Beam & save, crummy, slob & 0.220720\\ \hline
Random Attack + POS + Perplexity + Beam & next worthless misbegotten & 0.090090\\ \hline
\textbf{UAT Attack + POS + Perplexity + Beam} & \textbf{uselessly idiotic teleprompter, {irredeemably disgusting garbage}} & \textbf{0.040540} \\ \hline
\end{tabular}\\[1ex]
\caption{\label{positive_to_negative}Flipping positive to negative on the SST dataset. We present accuracy after the adversarial attack for different ablation experiments, and the corresponding generated trigger. Bolded experiment indicating our proposed method gives the lowest accuracy with sensible triggers.}
\end{table*}

\begin{table*}[h!]
\centering
\begin{tabular}{|p{6.7cm}|p{6.3cm}|p{1.5cm}|}
\hline
\textbf{Ablation Experiments} & \textbf{Sample Triggers} & \textbf{Accuracy} \\ \hline
Random Attack w/o Beam & succeeds absorbing courageousness & 0.212616\\ \hline
UAT Attack w/o Beam & powerfully edifying restored & 0.158878\\ \hline
Random Attack + POS + Beam & so enlightening companionship & 0.348130\\ \hline
Random Attack + POS + Perplexity + Beam & providing exuberant vitality & 0.271028\\ \hline
\textbf{UAT Attack + POS + Perplexity + Beam }& \textbf{fearlessly captivating vulnerability, wondrously radiant vitality} & \textbf{0.121495} \\ \hline
\end{tabular} \\[1ex]
\caption{\label{negative_to_positive}Flipping negative to positive on the SST dataset. We present accuracy after the adversarial attack for different ablation experiments, and the corresponding generated trigger. Bolded experiment indicating our proposed method gives the lowest accuracy with sensible triggers.}
\end{table*}

The extensions we consider in this project are sensible trigger generation and adversarial defense. 
\subsubsection{Sensible Trigger Generation} 
Although `universal' attacks are simpler to utilize, we observe that these methods often produce meaningless and ungrammatical text like "zoning tapping fiennes" \cite{wallace-etal-2019-universal}. Even an ordinary human wouldn't understand such nonsensical text, and hence this calls for producing meaningful trigger sequences that can still confuse the intended network. In our first extension, we propose a logic for generating sensible triggers for adversarial attacks. 

For the first extension to generate sensible triggers, we choose the task of sentiment analysis, the SST dataset and a trigger sequence length of 3 to compare results with strong baselines like \cite{wallace-etal-2019-universal} that report results with similar setup.

Our algorithm consists of the following components:

\begin{enumerate}
    \item \textbf{Trigger Generation Algorithm:} In terms of trigger generation, we use the algorithm specified in \cite{wallace-etal-2019-universal} without any changes. The algorithm generates \textit{num\_candidates} choices for each index in the trigger sequence, where \textit{num\_candidates} is a hyperparameter. We use \textit{num\_candidates} = 100.
    
    \item \textbf{POS Filtering:} Once we have the generated candidates for each index in the trigger, we only filter the tokens that match commonly occurring POS patterns of length 3 like \code{["ADV","ADJ","NOUN"], ["PRON", "VERB", "PRON"], ["ADV", "VERB", "PRON"], ["NOUN", "VERB", "ADJ"], ["VERB", "PRON", "VERB"], ["VERB", "PRON", "ADJ"], ["VERB", "PRON", "NOUN"]}. We use NLTK library to get the POS tags. The tags are from the universal tagset.
    
    \item \textbf{Beam Search:} We progressively find the trigger words token by token as outlined in \cite{wallace-etal-2019-universal}. Constraining the algorithm to choose only the best filtered token limits our desire to generate grammatical sequences, as it would give nonsensical jargon that facilitates the best attack. Thus, we use beam search to progressively generate tokens. We use a beam size of 5. 
    
    \item \textbf{Modified Loss Function:} We use scores from the GPT-2 language model to filter out ungrammatical sequences. We do this by using a modified loss function that weighs (with hyperparameters $\lambda$ and $\beta$) both the model loss and the perplexity of the trigger sequence from the language model. Contrary to the common meaning of loss, we want to maximize our loss that intends to maximize the classifier loss (cross-entropy loss) and minimize the perplexity of the generated sequence. We use a $\lambda$ of -0.00005 and $\beta$ of 5. Observing that the classifier loss was in the range 0-5 and perplexity in the range 0-100,000, we convert the perplexity values to the similar scale of classifier loss.
    
    \begin{equation}
    \text{loss} = \text{classifier\_loss} + \lambda \cdot \text{perplexity} + \beta
    \end{equation}
    
\end{enumerate}
The bolded experiment in tables \ref{positive_to_negative} and \ref{negative_to_positive} is our proposed approach for the extension. Without triggers, the task of converting positive to negative and negative to positive sentiment has an accuracy of 0.909909 and 0.892523 respectively. For each experiment, we present the sample trigger that gives the least accuracy on our task. Tables \ref{positive_to_negative} and \ref{negative_to_positive} show the different ablation experiments to illustrate why all components of the proposed method are required to produce the best results. The triggers show that it is possible to achieve low accuracies by producing grammatical triggers.

\subsubsection{Adversarial Defense}
\begin{figure}[h!]
    \centering
    \includegraphics[width=0.4\textwidth]{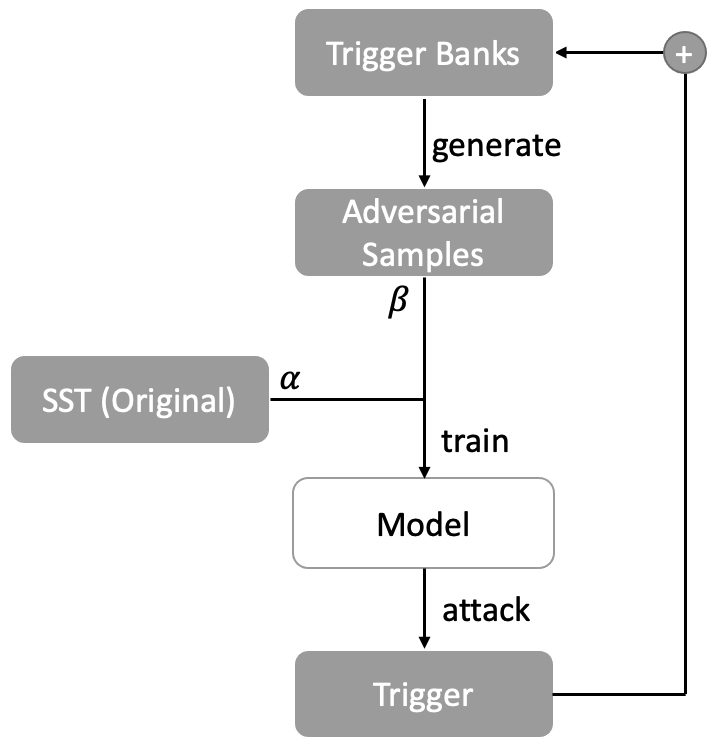}
    \caption{Adversarial Defense: Improving robustness by fine-tuning with adversarial samples. After training the model with the original dataset (SST), we iteratively generate the attack triggers and retrain the model by randomly augmenting the training set with the generated triggers.}
    \label{fig:defense_pipeline}
\end{figure}
In this section, we explore the possibility of improving the model robustness by utilizing the adversarial samples. Prior work \cite{liu-etal-2019-inoculation} used a challenge dataset to reveal and fix model weakness. In our approach, instead of using a pre-existing adversarial dataset as in prior work \cite{liu-etal-2019-inoculation}, we generate adversarial samples on the fly (at the end of each training iteration) and include them as part of the training data for the next iteration. In this way, the model can autonomously find weaknesses and train itself to defend against them.

Our self-supervision model pipeline is shown in figure \ref{fig:defense_pipeline}, and it has three important hyper-parameters:
\begin{enumerate}
    \item $\alpha \in \{0, 1\}$: Indicator function that indicates whether we include the original SST dataset in subsequent training iterations (fine-tuning).
    \item $\beta \in [0, 1]$: The percentage of the original dataset to which we randomly prepend triggers from the set of generated triggers.
    \item Number of epochs to train for each fine-tuning iteration. 
\end{enumerate}



\begin{figure*}[h!]
    \centering
    \includegraphics[width=0.8\textwidth]{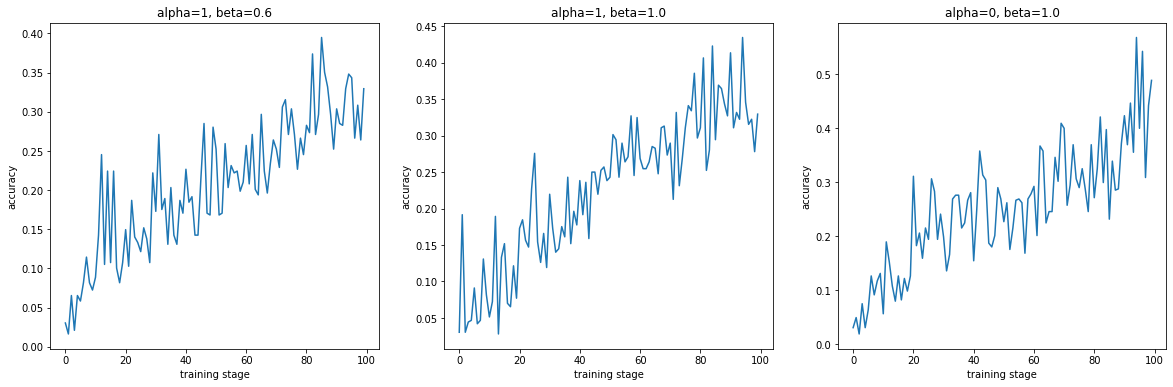}
    \caption{Model accuracy under attack (with triggers) v.s. number of iterations. (a) Left: fine-tuning with original SST dataset + 60\% of augmented adversarial samples (b) Middle: fine-tuning with original SST dataset + 100\% of augmented adversarial samples (c) Right: fine-tuning augmented adversarial samples only. We observe that fine-tuning with adversarial samples effectively improves the model. }
    \label{fig:defense_result_100}
\end{figure*}

\begin{table*}[h!]
\centering
\begin{tabular}{|c|c|c|}
\hline
                        & Accuracy on Original Dev Set & Accuracy on Attacked Dev Set \\ \hline
Baseline                & 0.8247                       & 0.0304                     \\ \hline
Defense: $\alpha = 1, \beta=0.6$ & 0.7874                       & 0.3294                     \\ \hline
Defense: $\alpha = 1, \beta=1.0$ & 0.7803                       & 0.3294                     \\ \hline
Defense: $\alpha = 0, \beta=1.0$ & \textbf{0.8388}              & \textbf{0.4883}           \\ \hline
\end{tabular}
\caption{Quantitative results of Adversarial Defense (fine-tuning). Attack Type: Flipping negative sentiment analysis to positive. Random seeds for torch, random and numpy were 52, 15, 43. }
\label{tab:defense_100}
\end{table*}
Table \ref{tab:defense_tuning_table} in the appendix shows the results from tuning hyperparameters over 20 iterations. We observe that all tested parameter sets result in a general upward trend in post-attack accuracy. Although there is significant fluctuation from iteration to iteration, the lowest outliers are still significantly above the levels before augmented training. With 20 iterations, there does not seem to be a significant difference between models trained on different numbers of epochs. Increasing the beta level, however, does seem to benefit performance, yielding a more pronounced upward trend in accuracy. (Note: We also included the training plots in the appendix section)

Figure \ref{fig:defense_result_100} and table \ref{tab:defense_100} show our final model performance. We terminated the fine-tuning after 100 attack-defense iterations (around 2 hours). Each attack-defense iteration involves training for a maximum of 5 epochs, which allows sufficient time for the model to converge. We observe that the overall trend for accuracy on attack dev set during the 100 iterations is monotonically increasing. In other words, the model becomes increasingly robust to attack as the training progresses (adding in more adversarial triggers). 

Surprisingly, fine-tuning with adversarial augmented data only (excluding original dataset) gives us the best performance among all attempts. We achieve over 83\% on the original dev set and over 48\% of accuracy under attack dev set. This probably suggests adversarial samples can effectively increase the diversity of training data, which in turn leads to better performance and robustness at the same time. This also indicates the vulnerability of the original model came from a dataset problem rather than a model problem.

Note that we also observe a certain amount of randomness in both training and evaluation. A possible cause for this may be the limited size of the SST dataset, which could lead to model over-fitting on random features.

\subsection{Error Analysis}
\subsubsection{Sensible Trigger Generation}
\begin{table*}[h!]
\centering
\begin{tabular}{|p{13cm}|}
\hline \textbf{Examples}  \\ \hline
`like leon , it 's frustrating and still oddly likable .'
 \\ \hline
 `corny , schmaltzy and predictable , but still manages to be kind of heartwarming , nonetheless .' \\
 \hline
`hardly a masterpiece , but it introduces viewers to a good charitable enterprise and some interesting real people .'\\ \hline
`first-time writer-director serry shows a remarkable gift for storytelling with this moving , effective little film .' \\
\hline
`macdowell , whose @@UNKNOWN@@ southern charm has anchored lighter @@UNKNOWN@@ ... brings an absolutely riveting conviction to her role .' \\ \hline
\end{tabular}
\caption{\label{sensible-error-analysis} Error analysis for sensible trigger generation. The table has a list of positive movie reviews that the generated sensible trigger "irredeemably disgusting garbage" is not able to flip.}
\end{table*}

Table \ref{sensible-error-analysis} shows some of the input movie reviews that our triggers fail to flip from positive to negative sentiment. We suspect that the trigger fails to flip these inputs for two possible reasons. First, if an input is overwhelmingly positive, the triggers may not sway the model enough to change its decision. A second possibility lies in the structure of the inputs. Some inputs start off by mentioning negative aspects but come to an overall positive conclusion, such as in the first two examples in the table. We suspect the model may be learning to recognize structures similar to subordinate clauses, where the latter part of a sentence has more influence on the overall sentiment. It is possible that when the input can be split by words such as 'but' or 'still,' the model favors the second half when deciding sentiment. 

\subsubsection{Adversarial Defense}
\begin{table*}[h!]
\centering
\begin{tabular}{|p{13cm}|}
\hline \textbf{Examples}  \\ \hline
 `the lower your expectations , the more you 'll enjoy it .' \\
 \hline
`you wo n't like roger , but you will quickly recognize him .'
\\ \hline
`this film seems for reflection , itself taking on adolescent qualities .' \\
\hline
\end{tabular}
\caption{\label{defense-error-analysis-1} Error analysis for adversarial defense. The table has a list of negative movie reviews where both the original model and the adversarially trained model succumb to adversarial attacks.}
\end{table*}

\begin{table*}[h!]
\centering
\begin{tabular}{|p{13cm}|}
\hline \textbf{Examples}  \\ \hline
`teen movies have really hit the @@UNKNOWN@@ .'
 \\ \hline
`it has all the excitement of eating @@UNKNOWN@@ .
' \\
\hline
`a great ensemble cast ca n't lift this heartfelt enterprise out of the familiar .
' \\ \hline
\end{tabular}
\caption{\label{defense-error-analysis-2} Error analysis for adversarial defense. The table has a list of negative movie reviews that the original model errs on, but the adversarially trained model resists attacks.}
\end{table*}

Tables \ref{defense-error-analysis-1} and \ref{defense-error-analysis-2} analyze errors of both the original model and the adversarially trained model. We see that the adversarially trained model resists trigger attacks better than the original model. Analyzing the common errors and the exclusive errors made by either of the models, we see that the errors are quite diverse and do not follow a specific pattern.

\section{Conclusions}
We developed a novel method for producing input-agnostic natural and grammatical triggers for attacking text classification models, giving better results than our strong baseline. We also performed experiments to increase the robustness of the models through adversarial training. Our aim of this work is to illustrate that adversarial attacks can be made harder to diagnose by generating grammatical triggers, and to develop robust NLP models through appropriate defenses.

\section{Acknowledgements}
We sincerely thank our project mentor Veronica Qing Lyu for her constant guidance throughout the course of this project work. 

Also, we thank Eric Wallace for his open-source implementation of \cite{wallace-etal-2019-universal} on Github that acted as our base code. \footnote{https://github.com/Eric-Wallace/universal-triggers}

\bibliographystyle{acl_natbib}
\bibliography{bibfile}
\appendix

\section{Additional Results}
\subsection{Baseline Results}
\begin{figure}[h!]
    \centering
    \includegraphics[width=0.4\textwidth]{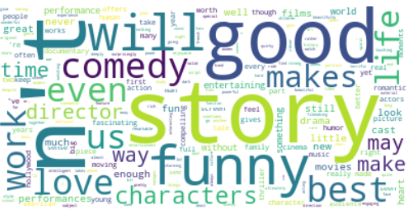}
    \caption{Top Frequency Words - Positive Sentiment }
    \label{fig:top_frequent_wordcloud_1}
\end{figure}

\begin{figure}[h!]
    \centering
    \includegraphics[width=0.4\textwidth]{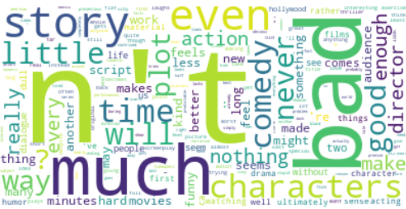}
    \caption{Top Frequency Words - Negative Sentiment }
    \label{fig:top_frequent_wordcloud_2}
\end{figure}

We use the SNLI dataset for this task. Tables \ref{entailment-1}, \ref{entailment-2}, \ref{contradiction-1}, \ref{contradiction-2}, \ref{neutral-1} and \ref{neutral-2} present the baseline results for different label pairs. Figures \ref{fig:top_frequent_wordcloud_1} and \ref{fig:top_frequent_wordcloud_2} show the top frequent words in a word cloud format (one of the simple baselines).

\begin{figure*}[h!]
    \centering
    \includegraphics[width=1.0\textwidth]{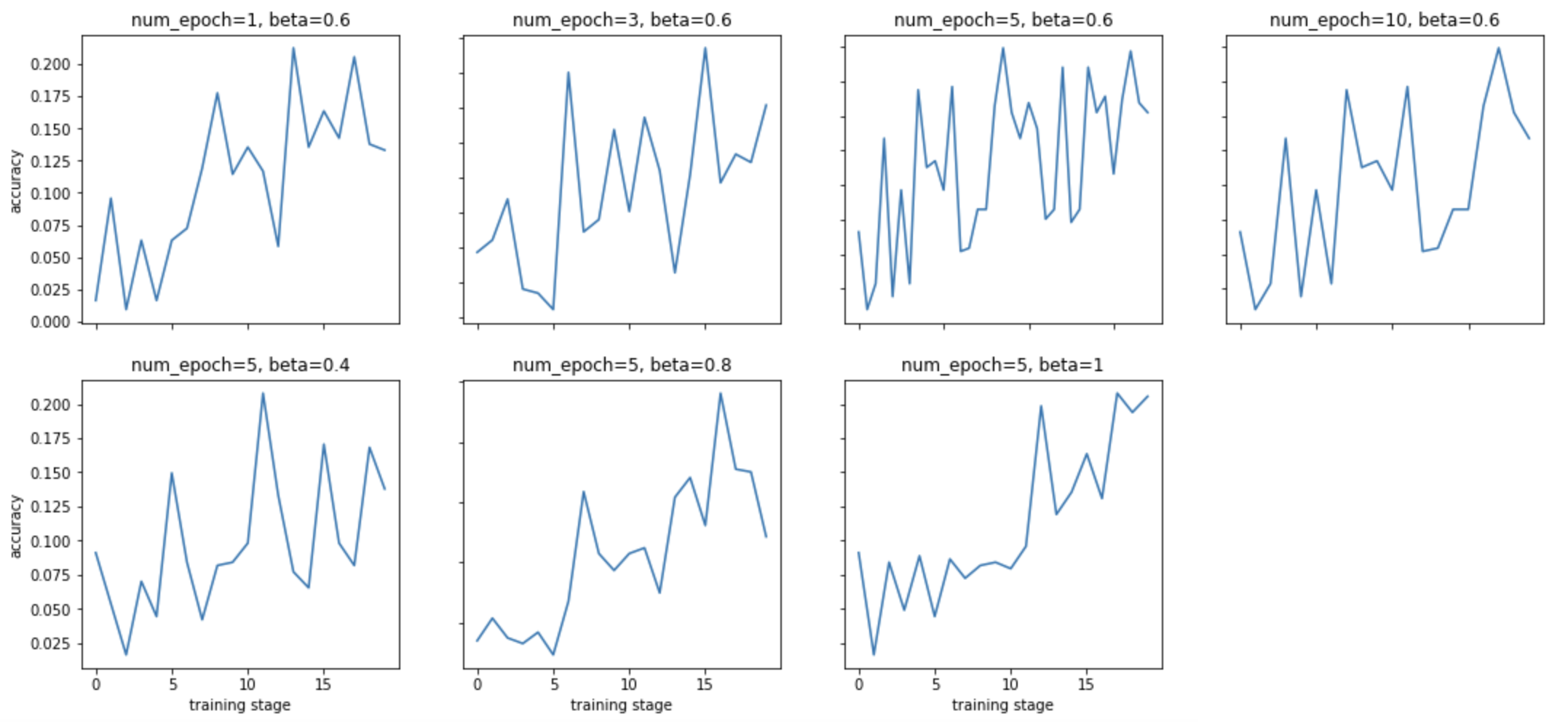}
    \caption{Model accuracy under attack (with triggers) v.s. number of iterations trained on different hyperparameters with 20 training stages. }
    \label{fig:defense_tuning_figure}
\end{figure*}

\begin{table*}[h!]
\centering
\begin{tabular}{|l |c | c | c |}
\hline
\textbf{Baseline} & \textbf{Trigger Used/Generated} & \textbf{Acc w/o trig} & \textbf{Acc with trig} \\ \hline
Random Triggers & sisters & 0.909582  & 0.006007 \\ \hline
Hardcoded Trigger & Except & 0.909582  & 0.051366 \\ \hline
Nearest Neighbor Trigger & nobody & 0.909582 & 0.00060  \\ \hline
Universal Adversarial Trigger \cite{wallace-etal-2019-universal} & mars & 0.909582 & 0.001201 \\ \hline
\end{tabular}
\caption{\label{entailment-1}Flipping entailment to contradiction}
\end{table*}

\begin{table*}[h!]
\centering
\begin{tabular}{|l |c | c | c |}
\hline
\textbf{Baseline} & \textbf{Trigger Used/Generated} & \textbf{Acc w/o trig} & \textbf{Acc with trig} \\ \hline
Random Triggers & zombie & 0.909582  & 0.001802 \\ \hline
Hardcoded Trigger & spaghetti & 0.909582  & 0.195854 \\ \hline
Nearest Neighbor Trigger &  no & 0.909582  & 0.001802 \\ \hline
Universal Adversarial Trigger \cite{wallace-etal-2019-universal} & joyously & 0.909582  & 0.000030  \\ \hline
\end{tabular}\\[1ex]
\caption{\label{entailment-2}Flipping entailment to neutral}
\end{table*}

\begin{table*}[h!]
\centering
\begin{tabular}{|l |c | c | c |}
\hline
\textbf{Baseline} & \textbf{Trigger Used/Generated} & \textbf{Acc w/o trig} & \textbf{Acc with trig} \\ \hline
Random Triggers & talents & 0.795302  & 0.667175 \\ \hline
Hardcoded Trigger & because &  0.795302 &  0.696156 \\ \hline
Nearest Neighbor Trigger & touching & 0.795302  & 0.666870 \\ \hline
Universal Adversarial Trigger \cite{wallace-etal-2019-universal} & amusing & 0.795302  & 0.660768 \\ \hline
\end{tabular}
\caption{\label{contradiction-1}Flipping contradiction to entailment}
\end{table*}

\begin{table*}[h!]
\centering
\begin{tabular}{|l |c | c | c |}
\hline
\textbf{Baseline} & \textbf{Trigger Used/Generated} & \textbf{Acc w/o trig} & \textbf{Acc with trig} \\ \hline
Random Triggers & festival &  0.795302  &  0.660768 \\ \hline
Hardcoded Trigger & spaghetti & 0.795302  & 0.897498 \\ \hline
Nearest Neighbor Trigger & anxiously & 0.795302  & 0.660158 \\ \hline
Universal Adversarial Trigger \cite{wallace-etal-2019-universal} &  joyously & 0.795302  & 0.595485 \\ \hline
\end{tabular}
\caption{\label{contradiction-2}Flipping contradiction to neutral}
\end{table*}

\begin{table*}[h!]
\centering
\begin{tabular}{|l |c | c | c |}
\hline
\textbf{Baseline} & \textbf{Trigger Used/Generated} & \textbf{Acc w/o trig} & \textbf{Acc with trig} \\ \hline
Random Triggers & rats & 0.880680  & 0.103554 \\ \hline
Hardcoded Trigger & because & 0.880680  & 0.956723 \\ \hline
Nearest Neighbor Trigger & mars & 0.880680  & 0.014219  \\ \hline
Universal Adversarial Trigger \cite{wallace-etal-2019-universal} & mars & 0.880680  & 0.014219 \\ \hline
\end{tabular}
\caption{\label{neutral-1}Flipping neutral to entailment}
\end{table*}

\begin{table*}[h!]
\centering
\begin{tabular}{|l |c | c | c |}
\hline
\textbf{Baseline} & \textbf{Trigger Used/Generated} & \textbf{Acc w/o trig} & \textbf{Acc with trig} \\ \hline
Random Triggers & no &  0.880680 & 0.035239 \\ \hline
Hardcoded Trigger & except & 0.880680  & 0.412364 \\ \hline
Nearest Neighbor Trigger & cat & 0.880680  & 0.005106 \\ \hline
Universal Adversarial Trigger \cite{wallace-etal-2019-universal} & cats &  0.880680 & 0.026275 \\ \hline
\end{tabular}
\caption{\label{neutral-2}Flipping neutral to contradiction}
\end{table*} 

\begin{table*}[h!]
\centering
    \begin{tabular}{|c|c|c|}
    \hline
     & Acc on Original Dev Set & Acc on Attacked Dev Set \\ \hline
     Baseline & 0.8247 & 0.0304\\ \hline
     $n$=1, $\beta$=0.6 & 0.8387 & 0.1144\\ \hline
     $n$=3, $\beta$=0.6 & 0.8154 & 0.1448\\ \hline
     $n$=5, $\beta$=0.4 & 0.8200 & 0.2102\\ \hline
     $n$=5, $\beta$=0.6 & \textbf{0.8294} & \textbf{0.2313} \\ \hline
     $n$=5, $\beta$=0.8 & 0.8411 & 0.1565\\ \hline
     $n$=5, $\beta$=1.0 & \textbf{0.7873} & \textbf{0.1705}\\ \hline
     $n$=10, $\beta$=0.6 & 0.8294 & 0.2313\\ \hline
    \end{tabular}
    \caption{Model accuracy under attack (with triggers) v.s. number of iterations trained on different hyperparameters with 20 training stages.}
    \label{tab:defense_tuning_table}
\end{table*}

\label{sec:appendix}
\end{document}